\documentclass[letterpaper, 10 pt, conference]{ieeeconf}  

\IEEEoverridecommandlockouts                              

\usepackage{graphics} 
\usepackage{epsfig} 
\usepackage{mathptmx} 
\usepackage{times} 
\usepackage{amsmath} 
\usepackage{amssymb}  
\usepackage{graphicx}
\usepackage{booktabs}
\usepackage{multirow}
\usepackage{subcaption}

\begin{document}



\title{LiftFeat: 3D Geometry-Aware Local Feature Matching}
\author{Yepeng Liu$^{1}$$^{\dagger}$, Wenpeng Lai$^{2}$$^{\dagger}$, Zhou Zhao$^{3}$, Yuxuan Xiong$^{1}$, Jinchi Zhu$^{1}$, Jun Cheng$^{4}$, and Yongchao Xu$^{1}$$^*$
\thanks{$^{1}$School of Computer Science, Wuhan University, Wuhan, China; 
$^{2}$SF Technology, Shenzhen, China; $^{3}$School of Computer Science, Central China Normal University and the Hubei Engineering Research Center for Intelligent Detection and Identification of Complex Parts, Wuhan, China; $^{4}$Institute for Infocomm Research, A*STAR, Singapore. ($^{\dagger}$: Equal contribution.) (Corresponding author: Yongchao Xu, yongchao.xu@whu.edu.cn)}
}

\maketitle
\begin{abstract}
Robust and efficient local feature matching plays a crucial role in applications such as SLAM and visual localization for robotics. Despite great progress, it is still very challenging to extract robust and discriminative visual features in scenarios with drastic lighting changes, low texture areas, or repetitive patterns. In this paper, we propose a new lightweight network called \textit{LiftFeat}, which lifts the robustness of raw descriptor by aggregating 3D geometric feature. Specifically, we first adopt a pre-trained monocular depth estimation model to generate pseudo surface normal label, supervising the extraction of 3D geometric feature in terms of predicted surface normal. We then design a 3D geometry-aware feature lifting module to fuse surface normal feature with raw 2D descriptor feature. Integrating such 3D geometric feature enhances the discriminative ability of 2D feature description in extreme conditions. Extensive experimental results on relative pose estimation, homography estimation, 
and visual localization tasks, demonstrate that our LiftFeat outperforms some lightweight state-of-the-art methods. 
Code will be released at : https://github.com/lyp-deeplearning/LiftFeat.

\end{abstract}

\section{INTRODUCTION}
Local feature matching between images is critical for many core robotic tasks, including Structure from Motion (SfM)~\cite{schonberger2016structure, jiang2020efficient, zhao2024automated}, Simultaneous Localization and Mapping (SLAM)~\cite{davison2007monoslam, mur2017orb, chung2023orbeez, liu2023regformer}, and visual localization~\cite{sattler2016efficient, sarlin2019coarse, wang2021continual, yin2023isimloc}. In practical applications, there are some scenes with extreme conditions, such as significant variation of illumination, and the presence of textureless or repetitive patterns. In these extreme conditions, achieving reliable feature matching still remains a challenging task.

Traditional local feature matching methods typically involve three stages: keypoint detection, descriptor extraction, and feature matching. Early methods such as SIFT~\cite{lowe2004distinctive} and SURF~\cite{bay2006surf} propose well-designed handcrafted descriptors. 
During the feature matching stage, nearest neighbor matching is commonly employed to obtain the matching results. 

In recent years, deep learning-based feature matching methods have significantly improved the performance of traditional algorithms~\cite{detone2018superpoint, liu2024progressive}. Some studies have jointly trained keypoint prediction and descriptor extraction~\cite{dusmanu2019d2, revaud2019r2d2, tyszkiewicz2020disk}, which not only increases processing speed but also further optimizes matching performance. Additionally, other studies have introduced graph neural networks~\cite{sarlin2020superglue, lindenberger2023lightglue}, framing the feature matching task as an optimal transport problem, thereby effectively improving matching accuracy.

\begin{figure}[ht]
    \centering
\includegraphics[width=0.48\textwidth]{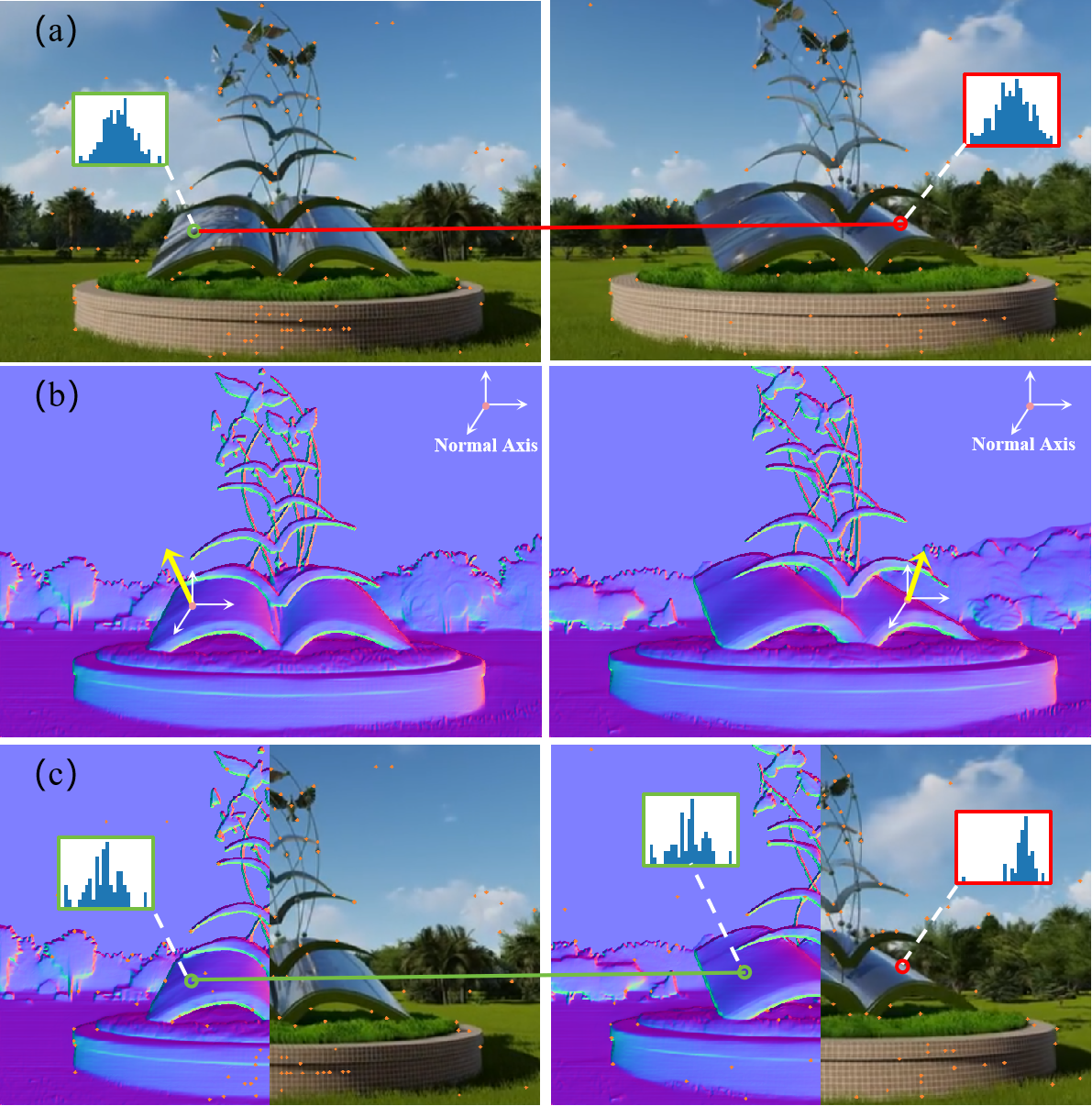}
    \caption{
 Feature matching of applying 2D visual cues and integrating 3D geometric cues in a low texture scene. Green lines: correct matches; Red lines: incorrect matches. Histograms: distribution of descriptor features.  {(a)} Result of using SuperPoint~\cite{detone2018superpoint}. {(b)} 3D geometric 
 Normal Map. {(c)} Result of using our LiftFeat. Incorporating 3D information enhances the distinctiveness of the raw 2D descriptors.
 }
\label{fig:motivate}
\end{figure}

Despite the advanced performance of current methods in most scenarios, 2D visual cues can cause confusion in feature matching for scenes with extreme conditions, including significant illumination variation, low texture, or repetitive patterns. 
As shown in Fig.~\ref{fig:motivate}, in textureless scenes, raw 2D descriptors may lead to incorrect matches due to insufficient discriminative visual information. An intuitive idea is to leverage the additional information from 3D data to enhance the robustness of feature matching. However, the precision and cost of using 3D data introduce new challenges, particularly in scenarios like robotics, where computational power is limited.

In this paper, we focus on designing a lightweight model that integrates 2D and 3D cues for local feature matching. Depth maps is one of the most accessible 3D cue. Yet,
depth maps exhibit scale ambiguity, making them unsuitable for direct use in local feature matching. In contrast, surface normal possesses both translation and scale invariance, which is suitable for feature matching. Therefore, we incorporate a surface normal estimation head into the network to learn 3D geometric knowledge.
Notably, the pseudo surface normal labels are derived from depth maps predicted by Depth Anything v2~\cite{yang2024depth}, which eliminates the need for additional annotation costs during training.
Subsequently, we propose a 3D Geometry-aware Feature Lifting (3D-GFL) module to fuse the raw 2D description with the 3D normal feature, lifting the discriminative ability of raw 2D descriptors in challenging scenarios. 
Experimental results demonstrate that our proposed method termed \textit{LiftFeat} achieves state-of-the-art performance across multiple tasks: relative pose estimation, homography estimation, and visual localization.

The main contributions of this work are as follows:
\begin{enumerate}
  \item  We propose a lightweight network named \textit{LiftFeat}, which innovatively introduces 3D geometry for local feature matching.
   \item We design a 3D Geometry-aware Feature Lifting (3D-GFL) module that fuses 2D description with 3D normal feature, significantly improving the discriminative ability of raw 2D descriptors in challenging scenarios. 
  \item Experiments on  different tasks confirm that our method achieves high accuracy and robustness across various scenarios. Additional runtime tests  confirm that our method can achieve  inference latency of 7.4 ms on edge devices.
\end{enumerate}

\section{related work}

\subsection{Local Feature Matching}
Local feature matching is a fundamental module in downstream applications such as visual localization~\cite{yin2023isimloc, liu2023nerf}, simultaneous localization and mapping (SLAM)~\cite{chung2023orbeez, adkins2024obvi}. It typically involves three key steps: feature detection, feature description, and feature matching. 
Traditional algorithms like SIFT~\cite{lowe2004distinctive} and ORB~\cite{rublee2011orb} focus on designing features that are invariant to scale, rotation, and illumination changes. Due to their ease of deployment, these methods are still widely used in robotics applications today.

With the development of deep learning, learning-based feature matching methods have achieved better matching performance. 
Some methods focus on jointly training keypoint detection and descriptor tasks~\cite{detone2018superpoint, revaud2019r2d2, zhao2022alike}, improving both efficiency and accuracy through multi-task optimization design.
In addition to keypoint detection and feature extraction, some methods have focused on improving feature matching performance. For instance, SuperGlue~\cite{sarlin2020superglue} and LightGlue~\cite{lindenberger2023lightglue} used graph neural networks (GNNs) and optimal transport optimization to effectively associate sparse local features while filtering outliers. 
However, these methods are not specifically designed for robotic platforms, and their inference time is relatively large.

Recently, some works have specifically designed lightweight networks for mobile VSLAM systems. 
Yao \textit{et al.}~\cite{yao2024edgepoint} designed a compact 32-dimensional descriptor using LocalPCA, enabling efficient storage and computation.
Su \textit{et al.}~\cite{su2024hpf} combined ORB and SuperPoint features, improving the accuracy in VSLAM systems.

Different with these methods, we introduce 3D geometric features while maintaining a lightweight design, enhancing the robustness of feature matching under extreme conditions in robotic applications.

\subsection{Feature Matching Leveraging 3D Information.}
3D features have been widely used in many downstream tasks~\cite{wofk2019fastdepth, xu2023onboard}, but their direct application in local feature matching has been relatively unexplored. In earlier study, Toft \textit{et al.}~\cite{toft2020single} improved image matching performance under large viewpoint changes by using features corrected through monocular depth estimation, but they did not directly leverage 3D features.  
Recently, Karpur \textit{et al.}~\cite{karpur2024lfm} introduced object spatial coordinate prediction in the object matching task and combined 3D coordinates with 2D features using an additional SuperGlue network.
Mao \textit{et al.}~\cite{mao20223dg} adopted a multi-modal training approach using a combination of depth maps and RGB inputs, enabling a dense matching network to learn implicit 3D features.
These methods have issues such as not directly utilizing 3D features and being highly time-consuming.

In this paper, we focus on designing a lightweight network that explicitly utilizes 3D features. We integrate surface normal prediction into the feature matching network, enhancing feature distinctiveness while maintaining efficiency.

\begin{figure*}[ht]
    \centering
\includegraphics[width=0.95\linewidth]{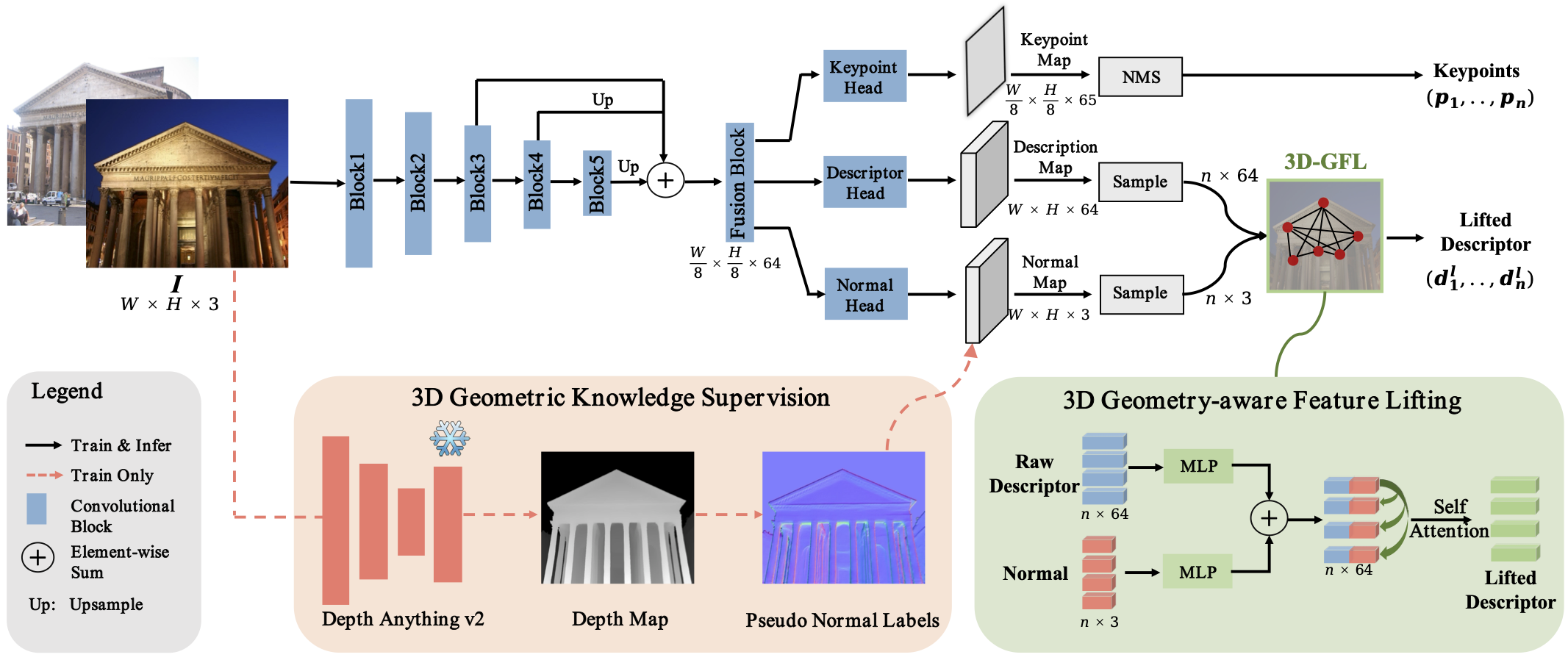}
    \caption{
    Overview of the proposed LiftFeat. Given an input image $I$, the feature extraction module outputs keypoint map, description map, and normal map through separate multi-task heads. During the training phase, we use the predicted depth map from the Depth Anything v2~\cite{yang2024depth} to obtain pseudo normal label as a supervisory signal to assist in learning 3D geometric features. Finally, the 3D geometric-aware feature lifting module fuses the 2D and 3D features.}
\label{fig:frame_work}
\end{figure*}

\section{method}
To address the limitations of   image feature matching in extreme scenarios, we propose a novel approach that leverages surface normal information from depth maps to enhance descriptor matching. In this section, we first present the network architecture of the proposed LiftFeat. Next, we explain how prior knowledge from a monocular depth estimation model is used to supervise the learning of surface normals. Furthermore, we introduce the 3D Geometry-aware Feature Lifting (3D-GFL) module that fuses surface normal information with original 2D descriptors. Finally, we introduce the network training details.

\subsection{Network Architecture} 
As shown in Fig.~\ref{fig:frame_work}, to achieve a better balance between accuracy and speed, we design a network architecture that consists of a shared feature encoding module and multiple task-specific heads. 

\medskip \noindent \textbf{Feature Encoding.}
Let the input image \( I \in \mathbb{R}^{W \times H \times 3} \), where \( W \) and \( H \) represent the width and height of the image, respectively. In the feature encoding module, we employ 5 blocks for feature extraction. All the blocks consist of \( 3 \times 3 \) convolution layers followed by max-pooling layers with a stride of 2. The output feature map from Block5 has a spatial resolution of \( \frac{W}{32} \times \frac{H}{32} \).
The depth of the feature maps increases progressively across the blocks, with the output depths of the 5 blocks being \{4, 8, 16, 32, 64\}, respectively. Subsequently, a fusion block performs multi-scale feature fusion on the lower-level features. We use \( 1 \times 1 \) convolutions and bilinear interpolation to align and sum the features from Block3, Block4, and Block5, resulting in a fused feature map of size \( \frac{W}{8} \times \frac{H}{8} \times 64 \).

\medskip \noindent \textbf{Multi-task Head.}
Our multi-task head is designed to predict keypoints, descriptors, and surface normals. The keypoint branch adopts a strategy similar to SuperPoint, where a \(1 \times 1\) convolution is applied to generate the keypoint map of size \(\frac{H}{8} \times \frac{W}{8} \times (64+1)\). A channel-wise softmax operation is then performed to obtain the keypoint score distribution at the original image resolution.
For the descriptor branch, we use bilinear interpolation and \(L_2\)-normalization operation to obtain a descriptor map of size \(W \times H \times 64\). Similarly, the normal head uses bilinear interpolation to obtain a 3-channel map with the same resolution as the original image.

\medskip \noindent \textbf{3D Geometry-aware Feature Lifting Module.} Based on the  keypoint information, we sample descriptors and normal features, which are then fed into 3D-GFL module to enhance the extracted features.

Assuming the keypoint branch predicts keypoints $p\in \mathbb{R}^{N \times 2}$ through Non-Maximum Suppression (NMS). We then perform a grid sample operation to extract the corresponding descriptor $d\in \mathbb{R}^{N \times 64}$ and normal vector features $n\in \mathbb{R}^{N\times 3}$.  Next, we align their feature dimensions by adding an Multi-Layer Perception (MLP) layer, followed by summing the aligned features. Finally, we apply stacked self-attention layers to obtain the lifted descriptors $d^l\in \mathbb{R}^{N\times 64}$.

\subsection{3D Geometric Knowledge Supervision}  
The surface normal describes the orientation of points on a surface and serves as a 3D signal with both translational and scale invariance. During the training phase, we utilize the monocular depth estimation model, Depth Anything v2~\cite{yang2024depth}, to generate supervision labels for the surface normals. Although monocular depth estimation inherently suffers from scale ambiguity, converting depth information into surface normals allows us to effectively mitigate this issue. This transformation enhances the performance of feature matching by providing robust geometric cues that are invariant to scale and translation.

Given an input image $I$, a depth estimation model is used to generate the corresponding depth map $Z_I$. 
For a pixel $P(u, v)$ in the image, its normal vector can be estimated based on the local gradient information. Let $Z_I(u, v)$ be the depth value at this point, and the depth gradients in the $u$ and $v$ directions can be approximated using finite differences as follows:

\begin{equation}
\frac{\partial Z_I}{\partial u} \approx Z_I(u+1, v) - Z_I(u-1, v),
\end{equation}
\begin{equation}
\frac{\partial Z_I}{\partial v} \approx Z_I(u, v+1) - Z_I(u, v-1).
\end{equation}

Using these depth gradients, we can estimate the normal vector at point $P(u, v)$, denoted as $\mathbf{n}_P$. Assuming the 3D coordinate of this point is $(u, v, Z_I(u,v))$, the normal vector $\mathbf{n}_P$ can be calculated as:

\begin{equation}
\mathbf{n}_P = \frac{(-\frac{\partial Z_I}{\partial u}, -\frac{\partial Z_I}{\partial v}, 1)}{\left\| \left( -\frac{\partial Z_I}{\partial u}, -\frac{\partial Z_I}{\partial v}, 1 \right) \right\|},
\end{equation}
where the denominator represents the magnitude of the vector, ensuring that the normal vector is normalized. The normal vector $\mathbf{n}_P$ provides information about the local surface orientation at point $P(u, v)$, reflecting the 3D geometric structure of the object.

\subsection{3D Geometry-aware Feature Lifting}
We integrate the 2D descriptors of keypoints with the 3D surface normal information using a feature aggregation module. Specifically, for each keypoint at coordinates $p_i$, we employ the grid sample operation to sample the corresponding local feature descriptor $d_i$ and the predicted surface normal vector $n_i$. 
Since the dimensions of the descriptors and  normals vector is different,  we employ separate multi-layer perceptron (MLP) layers to align the feature dimensions and perform the addition operation. Then, we use positional encoding (PE)~\cite{lindenberger2023lightglue, jiang2024Omniglue} to integrate the keypoint location information into the descriptor features, resulting in the mixed information $m_i$. The calculation process is as follows:
\begin{equation}
\mathbf{m}_i = PE (p_i) \odot ({MLP}_{2D}(\mathbf{d}_i) + {MLP}_{3D}(\mathbf{n}_i)).
\end{equation}

Following \cite{wang2022featurebooster}, we use stacked self-attention modules to enable the interaction and aggregation of feature information between different points. 
We use linear transformer layers to construct the self-attention module, which also enhances the model's inference speed.
For the $(n+1)^{th}$ layer, the corresponding feature $m_i^{n+1}\in \mathbb{R}^{D}$ ($D=64$ in this paper) of $i^{th}$ keypoint is aggregated from the original feature $m_i^{n}\in \mathbb{R}^{D}$ of the $n^{th}$ layer and the features of all other keypoints in $P$:
\begin{equation}
m_i^{n+1} = (m_i^{n}W_{m_i}^q) \odot \sum_{j \in P} \operatorname{Softmax}(m_j^{n}W_{m_j}^k) \odot (m_j^{n}W_{m_j}^v),
\label{eq:self attention}
\end{equation}
where $m_j$ represents the feature corresponding to a point in the set $P$.
$W_{m_i}^q$, $W_{m_j}^k$ and $W_{m_j}^v$ represent linear mapping layers. In the experiment, we use 3 self-attention layers.

\subsection{Network Training}
We supervise the network using pixel-level matching labels from paired images. The training data is derived from synthetic data or the Megadepth dataset~\cite{li2018megadepth}. Given a pair of input images $(I_A, I_B)$, we compute three types of losses: keypoint prediction loss $L_{keypoint}$, surface normal estimation loss $L_{normal}$, and descriptor loss $L_{desc}$.

\subsubsection{Keypoint Loss}
For keypoint supervision, we adopt  same strategy of SuperPoint~\cite{detone2018superpoint}. The original output of keypoint logits map is $(\frac{W}{8} \times \frac{H}{8} \times 65)$, where the last channel represents "no keypoint". We use the output of ALIKE detector~\cite{zhao2022alike} as the ground-truth keypoint labels. 
The keypoint loss $L_{keypoint}$ is computed by applying the Negative Log-Likelihood (NLL) loss on the keypoint logits map.

\subsubsection{Normal  Loss}
The normal vector estimation loss is applied to ensure accurate surface orientation predictions. For each predicted normal vector, we compare it with the ground-truth normal vector using the cosine similarity, ensuring that the predicted normal aligns with the true surface normal. The normal  loss is defined as:
\begin{equation}
L_{\text{normal}} = 1 - \frac{\mathbf{n}_{\text{pred}} \cdot \mathbf{n}_{\text{gt}}}{\|\mathbf{n}_{\text{pred}}\| \|\mathbf{n}_{\text{gt}}\|},
\end{equation}
where $\mathbf{n}_{\text{pred}}$ and $\mathbf{n}_{\text{gt}}$ are the predicted and ground-truth normal vectors, respectively.

\subsubsection{Descriptor Loss}

Given the descriptors sampled from image $I_A$ and $I_B$, we feed they to the feature fusion module to obtain the descriptors ($d_A \in \mathbb{R}^{m \times 64}$, $d_B \in \mathbb{R}^{n \times 64}$). Sequentially, we compute the similarity score matrix $S \in \mathbb{R}^{m \times n}$. The ground truth matching matrix is denoted as $M_{\text{gt}}$. Following ~\cite{sarlin2020superglue}, 
we minimize the negative log-likelihood of the predicted matching score matrix $S$ with respect to the ground-truth matching matrix $M_{\text{gt}}$:

\begin{equation}
L_{\text{desc}} = -\sum_{i,j} M_{\text{gt}}(i,j) \log S(i,j).
\end{equation}

\subsubsection{Total Loss}
The total loss for training is the weighted sum of these three components:
\begin{equation}
L_{\text{total}} =  L_{\text{keypoint}} + \alpha_1 L_{\text{normal}} + \alpha_2 L_{\text{desc}},
\end{equation}
where $\alpha_1$, and $\alpha_2$ are weighting factors that balance the contributions of the keypoint loss, normal loss, and descriptor loss, respectively. In this experiment, we empirically  set $\alpha_1$ and $\alpha_2$  to 2 and 1, respectively.

\begin{figure*}[t]
    \centering
\includegraphics[width=0.92\linewidth]{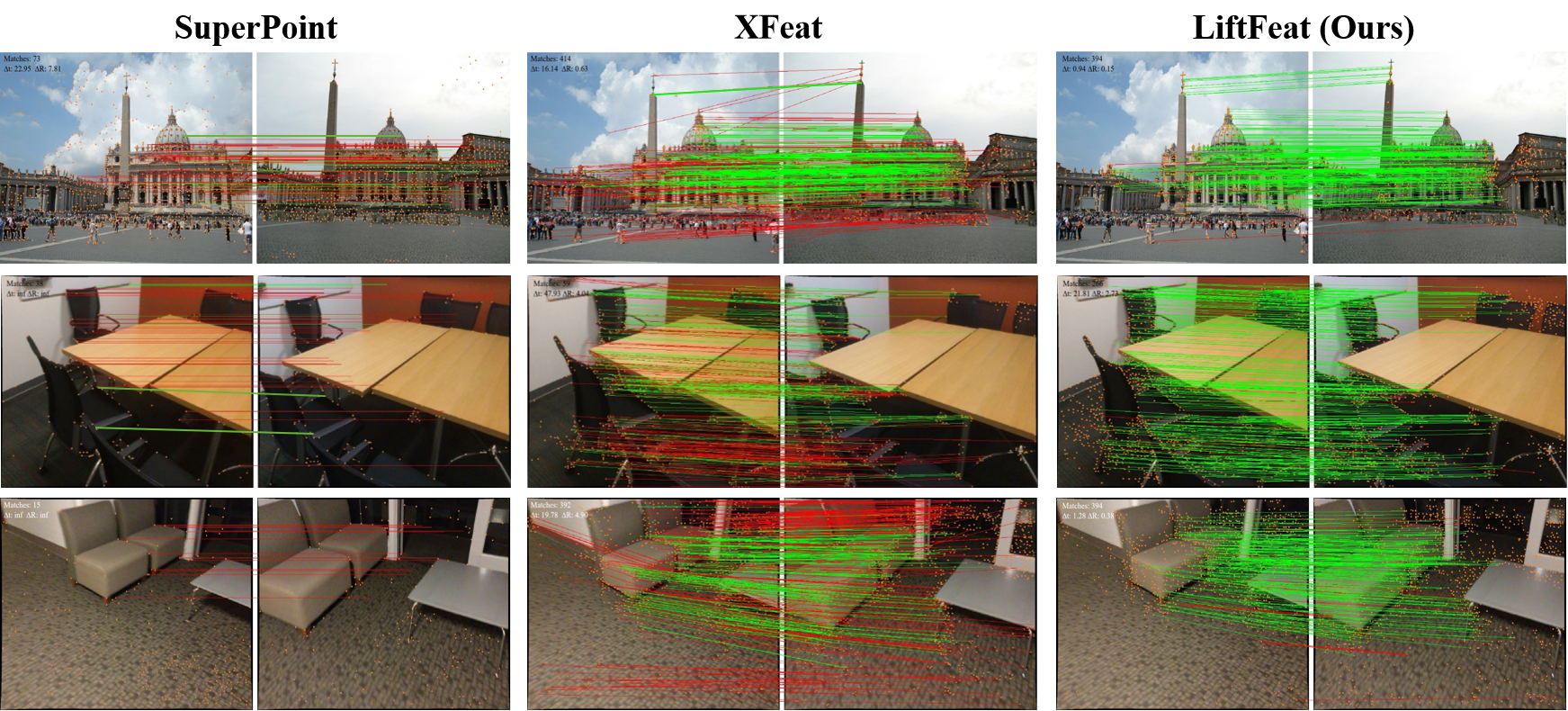}
    \caption{Qualitative matching results. We conduct tests in both indoor and outdoor scenes. The results demonstrate that our proposed LiftFeat maintains robust matching performance under extreme conditions, such as lighting variations (top), low texture (middle), and repetitive pattern (bottom) scenarios.  Green lines: correct matches; Red lines: incorrect matches.}
\label{fig:vis1}
\end{figure*}


\begin{table*}[t]
\centering
\caption{Relative pose results on MegaDepth-1500 and ScanNet. We report the AUC scores of translation and rotation errors at different thresholds. 
The best results are in bold, and the second-best are underlined.}
\begin{tabular}{lcccccccc}
\toprule
\textbf{Methods}  & \multicolumn{3}{c}{\textbf{MegaDepth-1500~\cite{li2018megadepth}}} & \multicolumn{3}{c}{\textbf{ScanNet~\cite{dai2017scannet}}} \\
 \cmidrule(lr){2-4} \cmidrule(lr){5-7}
 & AUC@5° & AUC@10° & AUC@20°  & AUC@5° & AUC@10° & AUC@20°    \\ 
\midrule
ORB~\cite{rublee2011orb}, ICCV2011 & 17.9 & 27.6 & 39.0 & 9.0 & 18.5 & 29.9  \\
SuperPoint~\cite{detone2018superpoint}, CVPRW2018 & 37.3 & 50.1 & 61.5 & 12.5 & 24.4 & 36.7  \\
ALIKE~\cite{zhao2022alike}, TIM2023 & 40.5 & \underline{56.9} & \underline{68.2} & 9.8 & 19.5 & 30.3 \\
SiLK~\cite{gleize2023silk}, ICCV2023 & 39.9 & 55.1 & 66.9 & 15.9 & 30.1 & 44.5  \\
XFeat~\cite{potje2024xfeat}, CVPR2024 & \underline{42.6} & 56.4 & 67.7 & \underline{16.7} & \underline{32.6} & \underline{47.8}  \\
LiftFeat (Ours)  & \textbf{44.7} & \textbf{59.5} & \textbf{70.3} & \textbf{18.5}  & \textbf{34.9} & \textbf{51.2}   \\
\bottomrule
\end{tabular}
\label{tab:relative pose estimate}
\end{table*}

\section{Experiments}
We evaluate the proposed LiftFeat on three tasks: relative  pose estimation, homography estimation and visual localization. The implementation details, comparative methods, and some qualitative illustrations are given in the following.


\medskip \noindent \textbf{Implementation Details.}
We implement the proposed algorithm based on PyTorch. During the training phase, we use the pre-trained Depth-Anything v2 model~\cite{yang2024depth} to generate pseudo surface normal. The training dataset is composed of a mixed dataset from MegaDepth~\cite{li2018megadepth} and synthetic COCO~\cite{lin2014microsoft}. 
The input image size is 800$\times$600 pixels. The model is optimized using the Adam optimizer with an initial learning rate of 1e-4 and a batch size of 16. During training, we sample 1024 pairs of matching points to fine-tune the feature aggregation module. The model training is completed in 32 hours on an NVIDIA RTX 3090. 

To verify the robustness of the method, we do not perform additional fine-tuning in all experiments.

\medskip \noindent \textbf{Comparative methods.}
Due to the computational limitations of the robot, we select some lightweight baseline methods: ORB~\cite{rublee2011orb}, SuperPoint~\cite{detone2018superpoint}, ALIKE~\cite{zhao2022alike}, SilK~\cite{gleize2023silk} and XFeat~\cite{potje2024xfeat}. For SiLK and ALIKE, we choose their smallest available backbones ALIKE-Tiny and VGG-aligning with our focus on computationally efficient models. For all baselines, we use the top 4096 detected keypoints. During matching, we employe mutual nearest neighbor (MNN) search.

\medskip \noindent \textbf{Qualitative illustrations.}
Fig.~\ref{fig:vis1} illustrates the visualization results in scenarios with low textures, repetitive patterns, and lighting variations. Our proposed LiftFeat enhances the discriminative ability of descriptors by incorporating 3D geometric features, improving the accuracy of feature matching in extreme  conditions.

\subsection{Relative Pose Estimation}
\medskip \noindent \textbf{Datasets.}
We evaluate our model on two commonly used datasets: MegaDepth-1500~\cite{li2018megadepth}  and ScanNet~\cite{dai2017scannet}. These images include challenging scenes with significant variations in viewpoint and lighting conditions. 
MegaDepth-1500~\cite{li2018megadepth} is an outdoor dataset containing multiple scenes. 
ScanNet~\cite{dai2017scannet} is an indoor RGB-D dataset consisting of 1613 sequences and 2.5 million views, each accompanied by ground-truth camera poses and depth maps. 
Following the setup of XFeat~\cite{potje2024xfeat}, the maximum size of MegaDepth~\cite{li2018megadepth} is 1200 pixels, while VGA resolution is used for testing images in ScanNet~\cite{dai2017scannet}.

\medskip \noindent \textbf{Metrics.}
Following ~\cite{lindenberger2023lightglue, potje2024xfeat}, we report the Area Under the recall Curve (AUC) for translation and rotation errors at various thresholds ($5^\circ$, $10^\circ$, $20^\circ$). The pose is computed using the essential matrix through the MAGSAC++~\cite{barath2020magsac++} algorithm.

\medskip \noindent \textbf{Results.}
As shown in Tab.~\ref{tab:relative pose estimate}, we present the results of pose estimation in both indoor and outdoor scenes. 
Compared to the newest lightweight network XFeat~\cite{potje2024xfeat}, we achieve significant improvements in AUC@5, AUC@10, and AUC@20 under the matching of 4096 sparse keypoints. Compared to SuperPoint~\cite{detone2018superpoint}, our approach demonstrates advantages in both accuracy and speed. This indicates that incorporating 3D geometric knowledge can significantly improve the accuracy of pose estimation.

\begin{table}[t]
\centering
\caption{Homography estimation results on HPatches~\cite{balntas2017hpatches}. We report  mean homography accuracy at different thresholds. 
The best are in bold, and the second-best are underlined.}
\begin{tabular}{lccc|ccc}
\toprule
\multirow{2}{*}{\textbf{Methods}} & \multicolumn{3}{c}{\textbf{Illumination}} & \multicolumn{3}{c}{\textbf{Viewpoint}} \\
\cmidrule(lr){2-4} \cmidrule(lr){5-7}
                        & @3  & @5  & @7  & @3  & @5  & @7  \\
\midrule
ORB~\cite{rublee2011orb}                     & 74.6  & 84.6  & 85.4  & 63.2  & 71.4  & 78.6  \\
SuperPoint~\cite{detone2018superpoint}              & {94.6}  & \underline{98.5}  & {98.8}  & \textbf{71.1}  & {79.6}  & {83.9}  \\
ALIKE~\cite{zhao2022alike}                   & {94.6}  & \underline{98.5}  & \textbf{99.6}  & 68.2  & 77.5  & 81.4  \\
SiLK~\cite{gleize2023silk}                    & 78.5  & 82.3  & 83.8  & 48.6  & 59.6  & 62.5  \\
XFeat~\cite{potje2024xfeat}                   & \underline{95.0}  & 98.1  & 98.8  & \underline{68.6}  & \underline{81.1}  & \underline{86.1}  \\
LiftFeat (Ours)                   & \textbf{95.6}  & \textbf{98.8} & \underline{99.2}  & \textbf{71.1}  & \textbf{81.7}  & \textbf{87.5} \\

\bottomrule
\end{tabular}
\label{tab:hpatch}
\end{table}

\begin{table}[t]
\centering
\caption{Visual localization on Aachen Day-Night~\cite{sattler2018benchmarking}. 
We report the pose recall at (0.25m/2°, 0.5m/5°, 5m/10°). 
The best  are in bold, and the second-best are underlined.}
\resizebox{\columnwidth}{!}{
\begin{tabular}{lcccccc}
\toprule
\multirow{2}{*}{\textbf{Methods}} & \multicolumn{3}{c}{\textbf{Day}} & \multicolumn{3}{c}{\textbf{Night}} \\
\cmidrule(lr){2-4} \cmidrule(lr){5-7}
 & 0.25m, 2° & 0.5m, 5° & 5m, 10° & 0.25m, 2° & 0.5m, 5° & 5m, 10° \\
\midrule
ORB~\cite{rublee2011orb} & 66.9 & 76.1 & 93.7 & 10.2 & 12.2 & 19.4 \\
SuperPoint & \underline{87.4} & \textbf{93.2} & \underline{97.0} & 77.6 & {85.7} & 95.9 \\
ALIKE & {85.7} & {92.4} & {96.7} & \underline{81.6} & {88.8} & \underline{99.0} \\
XFeat & {84.7} & 91.5 & 96.5 & {77.6} & \underline{89.8} & 98.0 \\
LiftFeat (Ours) & \textbf{87.6} & \underline{93.1} & \textbf{97.1} & \textbf{82.1} & \textbf{89.9} & \textbf{99.1} \\
\bottomrule
\end{tabular}
}
\label{tab:localization}
\end{table}

\subsection{Homography Estimation}
\medskip \noindent \textbf{Datasets.} 
We use the widely adopted HPatches~\cite{balntas2017hpatches} dataset to evaluate homography. HPatches~\cite{balntas2017hpatches} consists of planar sequences with various lighting and viewpoint changes. Each scene contains 5 image pairs, accompanied by ground truth homography matrices.

\medskip \noindent \textbf{Metrics.}
Following ~\cite{zhao2022alike}, we report the Mean Homography Accuracy (MHA) metric. The MHA measures the proportion of images where the average error between the mapped and ground truth corner points, calculated using the estimated homography matrix, falls within a  pixel threshold. In our experiments, we set different thresholds of $\{3, 5, 7\}$ pixels.

\medskip \noindent \textbf{Results.}
Tab.~\ref{tab:hpatch} shows the results of HPatches~\cite{balntas2017hpatches} under varying illumination and viewpoint conditions. Our method generally outperforms other algorithms. Particularly in scenarios with large viewpoint changes, geometric distortions can cause significant alterations in the appearance features on 2D images. Introducing 3D information can help mitigate this effect.

\subsection{Visual Localization}
\medskip \noindent \textbf{Datasets.} 
We demonstrate the performance of our approach on the Aachen Day-Night v1.1~\cite{sattler2018benchmarking} dataset for visual localization tasks. This dataset presents challenges in terms of illumination changes and contains 6,697 daytime database images along with 1,015 query images (824 captured during the day and 191 at night).
The ground truth 6DoF camera poses are obtained using COLMAP~\cite{schonberger2016structure}. During testing, we resize the images to a maximum dimension of 1024 pixels and extract the top 4096 keypoints from all methods.

\medskip \noindent \textbf{Metrics.}
We use the hierarchical localization toolbox (HLoc)~\cite{sarlin2019coarse} by replacing the feature extraction module with different feature detectors and descriptors. Then, we report the accuracy of correctly estimated camera poses within position error thresholds of {0.25m, 0.5m, 5m} and rotation error thresholds of {2°, 5°, 10°}.

\medskip \noindent \textbf{Results.}
Tab.~\ref{tab:localization} shows results of  visual localization. Our method outperforms  ALIKE~\cite{zhao2022alike} and XFeat~\cite{potje2024xfeat}, in both daytime and nighttime scenarios. Compared with the widely-used industrial algorithm SuperPoint~\cite{detone2018superpoint}, our performance in daytime scenarios is comparable. However, in nighttime scenarios, under the threshold of (0.25m/1°), we improve the success rate from 77.6\% to 82.4\%. This suggests that in nighttime scenes, 3D cues can generate more distinctive features under the same conditions.


\begin{table}[t]
\centering
\caption{Ablation study on visual localization task with night subset of Aachen Day-Night dataset~\cite{sattler2018benchmarking}. The default setting includes only keypoint and description prediction.}

\begin{tabular}{lccc}
\toprule
{\textbf{Methods}} & (0.25m, 2°) & (0.5m, 5°) & (5m, 10°) \\
\midrule
Default &  78.9 & 86.1 & 97.6 \\
+ Normal Head &  79.4 & 87.9 & 98.2 \\
+ 3D-GFL &   82.1 & 89.9 & 99.1 \\
\bottomrule
\end{tabular}
\label{tab:ablation}
\end{table}
\subsection{Ablation Study}
In this section, we analyze the impact of adding normal head to learn the 3D geometric knowledge and the  3D-GFL module. Our baseline setup only includes the keypoint detection and raw description prediction. 
We conduct the ablation study on a highly challenging nighttime visual localization test set.
From Tab.~\ref{tab:ablation}, it can be observed that adding multi-task heads in an implicit manner yields gains of (0.5\%, 1.8\%, 0.6\%). On this basis, explicit feature aggregation further improves the accuracy, achieving gains of (2.7\%, 2.0\%, 0.9\%).

\begin{table}[t]
\centering
\caption{Comparison of computation resources.}
\begin{tabular}{lccc}
\toprule
 & SuperPoint~\cite{detone2018superpoint} & XFeat~\cite{potje2024xfeat} & Ours \\ 
\midrule
{Params (M)} & 1.30 & 0.66 & 0.85 \\ 
{FLOPs (G)} & 19.85 & 1.33 & 4.96 \\ 
{Desc. Dimension} & 256-f & 64-f & 64-f \\ 
{runtime/CPU (ms)} & 227   & 35     & 62 \\ 
{runtime/GPU (ms)} &  36   & 5.6    & 7.4   \\ 

\bottomrule
\end{tabular}
\label{tab:runtime}
\end{table}

\subsection{Runtime Analyse}
We compare the resource requirements for deploying two widely used methods, SuperPoint~\cite{detone2018superpoint} and XFeat~\cite{potje2024xfeat}, on edge devices. For the CPU, we selecte an Intel(R) Core(TM) i7-10700 CPU @ 2.90GHz, and for the mobile GPU, we choose the commonly used Nvidia Xavier NX. 
We use feed the  real-time VGA data into the network.  As shown in Tab.~\ref{tab:runtime}, while our method is slightly slower than XFeat~\cite{potje2024xfeat}, it outperforms XFeat in accuracy across all three tasks. Compared to SuperPoint~\cite{detone2018superpoint}, our method is 5 times faster and also more accurate. This demonstrates that our approach achieves a good balance between accuracy and speed.

\section{CONCLUSIONS}
In this paper, we present a novel lightweight network for 3D geometry-aware local feature matching. We propose to learn surface normal for encoding the 3D geometric feature. For that, we leverage the depth anything model to estimate depth map, based on which we derive the pseudo surface normal for supervision. The proposed method termed \textit{LiftFeat} then effectively aggregates 3D geometry feature of learned surface normal into raw 2D description. This lifts the discrimination ability of visual feature, in particular for scenes with extreme conditions such as significant lighting changes, low textures, or repetitive patterns. 
The superiority over some lightweight state-of-the-art methods is validated on three tasks: relative pose estimation, homography estimation and visual localization. 

\medskip \noindent \textbf{Acknowledgment.}
This work was supported in part by the National Key Research and Development Program of China (2023YFC2705700), NSFC 62222112, and 62176186, the Innovative Research Group Project of Hubei Province under Grants (2024AFA017), the Postdoctoral Fellowship Program of CPSF (No. GZC20230924), the Open Projects funded by Hubei Engineering Research Center for Intelligent Detection and Identification of Complex Parts (No. IDICP-KF-2024-03), Hubei Provincial Natural Science Foundation (No. 2024AFB245), and the Agency for Science, Technology and Research under its MTC Programmatic Funds Grant No.M23L7b0021.









\newpage

\bibliographystyle{ieeetr}
\bibliography{root}

\end{document}